\pdfoutput=1

\documentclass[11pt]{article}

\usepackage{latex/acl}

\usepackage{times}
\usepackage{latexsym}

\usepackage[T1]{fontenc}

\usepackage[utf8]{inputenc}

\usepackage{microtype}

\usepackage{inconsolata}
\usepackage{xcolor}
\usepackage{colortbl}
\definecolor{dt}{HTML}{ADCAD8}
\definecolor{dt2}{HTML}{cddfe7}

\usepackage{graphicx} 
\usepackage{graphicx}
\usepackage{amsmath}
\usepackage{amssymb}
\usepackage{booktabs}
\usepackage{makecell}
\usepackage{multirow}
\usepackage{xspace}
\usepackage[justification=centering]{subcaption}
\newcommand{\modelname}{LaDiC\xspace}

\newcommand{\grayrow}[1]{\textcolor{gray}{#1}}
\newcommand*\samethanks[1][\value{footnote}]{\footnotemark[#1]}

\usepackage{algorithm}
\usepackage{algorithmic}
\usepackage{adjustbox}
\title{LaDiC: Are Diffusion Models Really Inferior to Autoregressive Counterparts for Image-to-Text Generation? }

\author{Yuchi Wang$^{1}$\thanks{~~Equal contribution.}, ~Shuhuai Ren$^{1}$\samethanks, ~Rundong Gao$^1$,~Linli Yao$^1$ ,~Qingyan Guo$^2$, \\ \textbf{Kaikai An$^1$,~Jianhong Bai$^3$,~Xu Sun$^1$\thanks{~~Corresponding author.}} \\
 $^{1}$ National Key Laboratory for Multimedia Information Processing, Peking University \\
   $^{2}$Tsinghua University
   $~^{3}$Zhejiang University \\
   \texttt{\{wangyuchi,shuhuai\_ren\}@stu.pku.edu.cn
   ~xusun@pku.edu.cn}
}

\begin{document}
\maketitle

\begin{abstract}
Diffusion models have exhibited remarkable capabilities in text-to-image generation. However, their performance in image-to-text generation, specifically image captioning, has lagged behind Auto-Regressive (AR) models, casting doubt on their applicability for such tasks. In this work, we revisit diffusion models, highlighting their capacity for holistic context modeling and parallel decoding. With these benefits, diffusion models can alleviate the inherent limitations of AR methods, including their slow inference speed, error propagation, and unidirectional constraints. Furthermore, we identify the prior underperformance of diffusion models stemming from the absence of an effective latent space for image-text alignment, and the discrepancy between continuous diffusion processes and discrete textual data. In response, we introduce a novel architecture, \modelname, which utilizes a split BERT to create a dedicated latent space for captions and integrates a regularization module to manage varying text lengths. Our framework also includes a diffuser for semantic image-to-text conversion and a Back$\&$Refine technique to enhance token interactivity during inference. \modelname achieves state-of-the-art performance for diffusion-based methods on the MS COCO dataset with 38.2 BLEU@4 and 126.2 CIDEr, demonstrating exceptional performance without pre-training or ancillary modules. This indicates strong competitiveness with  AR models, revealing the previously untapped potential of diffusion models in image-to-text generation.\footnote{Code released at \url{https://github.com/wangyuchi369/LaDiC}}

\end{abstract}
\section{Introduction} \label{sec: intro}

\begin{figure}[t!]
    \centering
\includegraphics[width=0.95\linewidth]{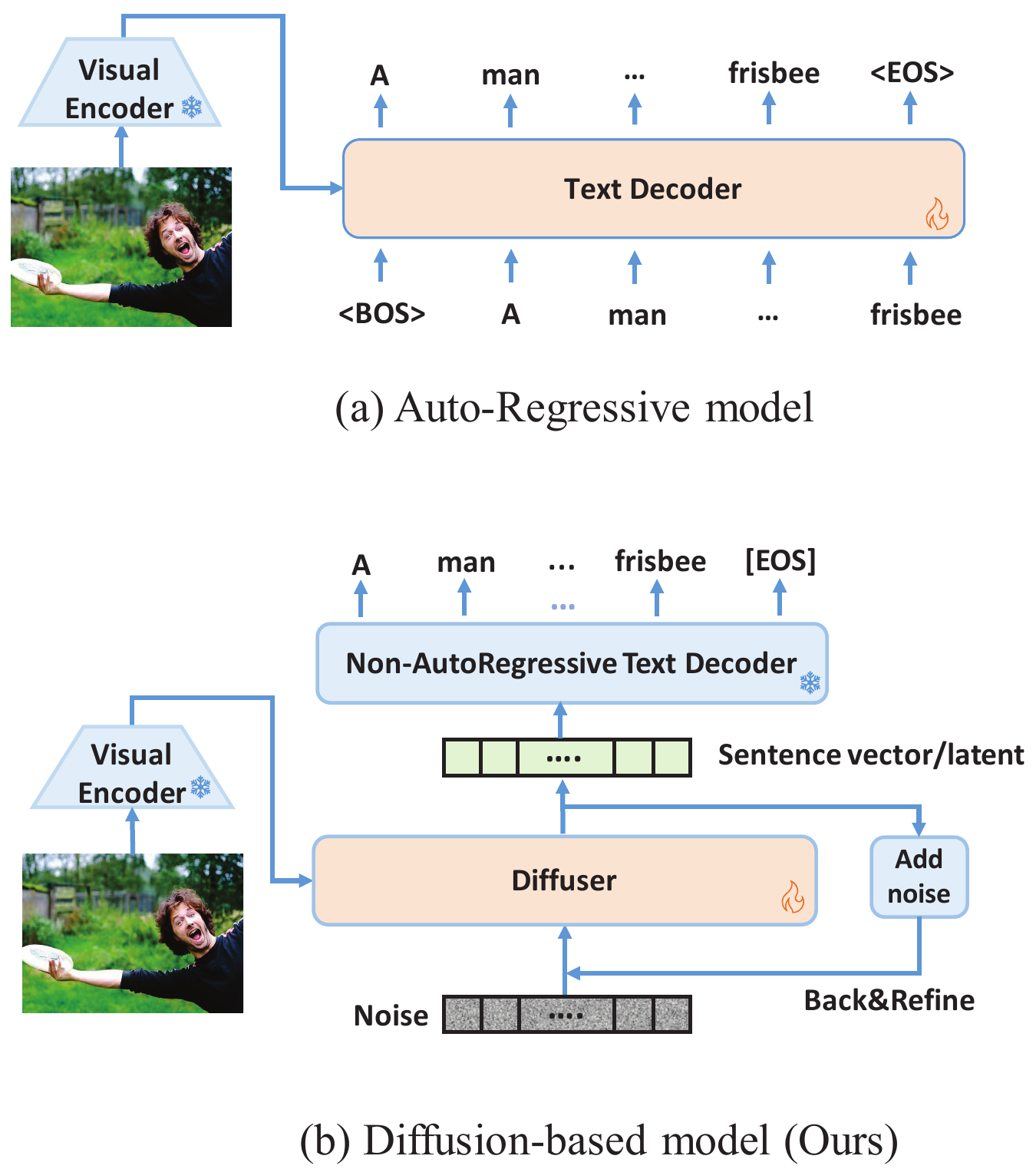}
    \caption{Inference process for image captioning. (a)  Token-by-token generation manner of AR-based model. (b)  Gradually denoising generation manner of the diffusion-based model (Ours). }
    \label{fig:intro}
\end{figure}

\begin{figure*}[htbp] %
    \centering
    

   \begin{minipage}[c]{0.9\textwidth}
        \centering
        \begin{subfigure}[c]{0.49\textwidth}
            \centering
            \includegraphics[width=1\textwidth]{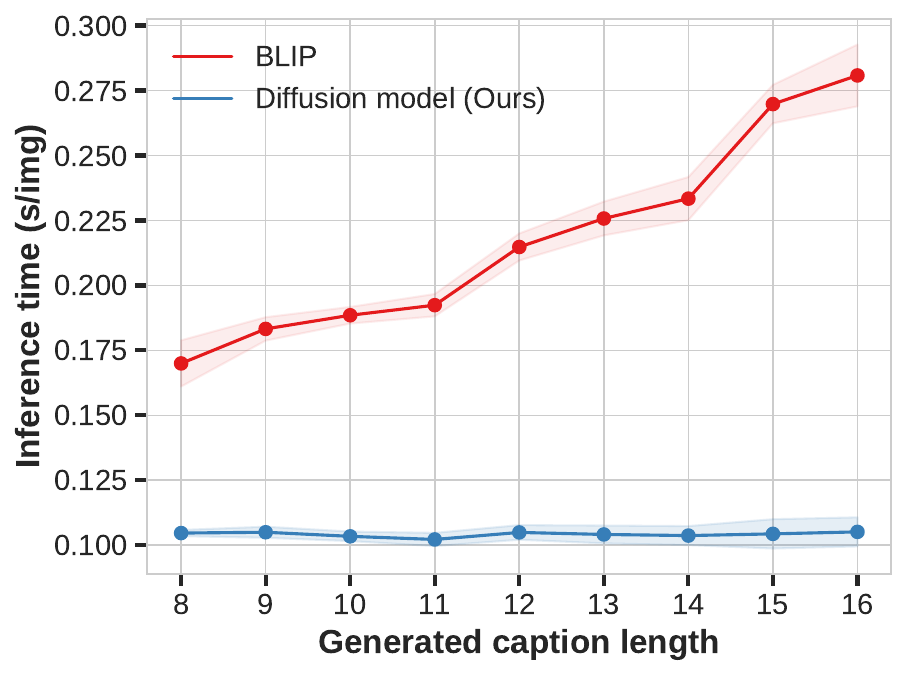}
            \caption{Inference time as sequence length increases.}
            \label{fig:subfig1}
        \end{subfigure}
        \hfill
        \begin{subfigure}[c]{0.49\textwidth}
            \centering
            \includegraphics[width=1\textwidth]{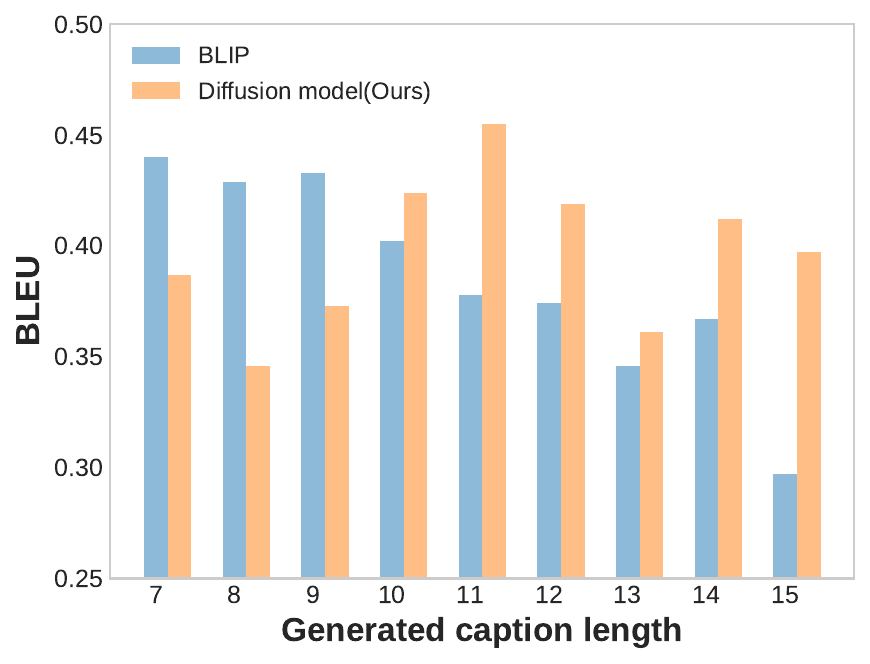}
            \caption{BLEU score as sequence length increases.}
            \label{fig:subfig2}
        \end{subfigure}
    \end{minipage}
    \vspace{0.1in}
    \begin{minipage}[c]{0.8\textwidth}
        \centering
        \begin{subfigure}[c]{\textwidth}
            \centering
            \includegraphics[width=\textwidth]{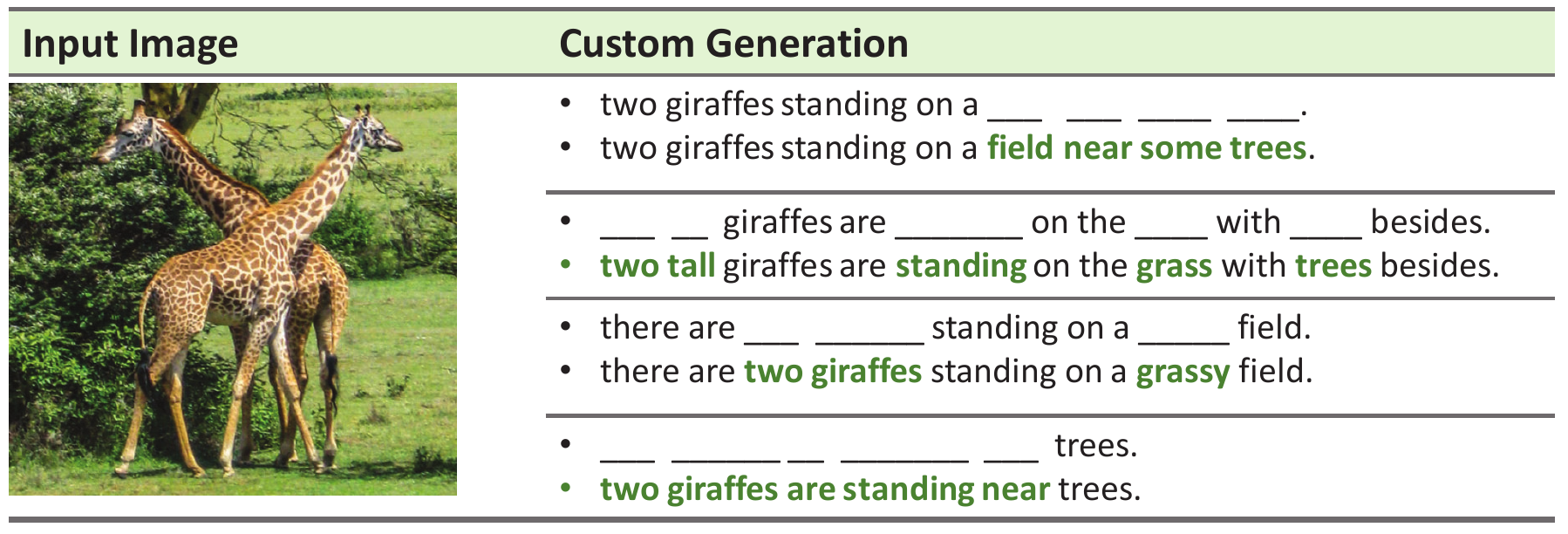}
            \caption{\modelname's ability of custom generation.}
            \label{fig:subfig3}
        \end{subfigure}
    \end{minipage}
    
    \caption{Three advantages of our diffusion-based model (\modelname) compared to auto-regressive models (BLIP).}
    \label{fig:adv}
\end{figure*}

In recent years, there has been a surge of impressive applications of diffusion models in text-to-image generation tasks~\citep{DALLE3, podell2023sdxl, dai2023emu}. 
However, the inverse process of image-to-text generation remains less explored. Some pioneering efforts~\citep{difflm, Yuan2022SeqDiffuSeqTD} have attempted to integrate diffusion models into text generation tasks. They mainly followed the traditional Encoder-Decoder framework, utilizing the diffusion model as a text decoder. Subsequent research~\citep{He2023DiffCapEC, Liu2023PrefixdiffusionAL} introduces visual capability into this paradigm by treating visual inputs as special tokens or encoded hidden states, thereby extending the research scope to the realm of multi-modal tasks, such as image-to-text generation. 
However, their performance has consistently lagged behind that of Auto-Regressive (AR) models~\citep{blip, vinvl, Wang2022GITAG}. Only through intricate architecture~\citep{scd} or external data~\citep{ddcap} can they barely achieve comparable results, raising doubts about whether diffusion models have inherent limitations, potentially making them less suitable for the image-to-text task.

In this study, we aim to dispel this doubt by deeply reexamining the diffusion-based image-to-text paradigm and unveiling its distinct benefits.  Unlike conventional AR approaches that sequentially generate captions token by token (Fig.~\ref{fig:intro}a), diffusion-based models take Gaussian noise as input and iteratively denoise it under image guidance to simultaneously produce the entire caption (Fig.~\ref{fig:intro}b). 
This Non-AutoRegressive (NAR) diffusion-based model exhibits three key advantages:
\textbf{(1) Parallel Decoding:} Diffusion-based models emit all tokens in parallel, significantly reducing inference time for lengthy target captions. As illustrated in Fig.~\ref{fig:subfig1}, the inference time of AR models like BLIP~\citep{blip} proliferates as text length increases, while our diffusion model can ensure stable inference time regardless of the length increase.  For instance, when the caption length reaches 16, our model is approximately 3$\times$ faster than BLIP.
~\textbf{(2) Holistic Context Consideration:} Unlike the uni-directional information flow of AR models (left to right), diffusion-based models can consider more holistic contexts, mitigating error accumulation~\citep{He2023DiffCapEC}. As depicted in Fig.~\ref{fig:subfig2}, the BLEU metric of BLIP-generated captions declines rapidly with increasing text length, whereas our diffusion-based model maintains performance.
\textbf{(3) Flexible Generation:} AR models adhere to a fixed unidirectional generation manner, whereas our diffusion model demonstrates much greater flexibility. As shown in Fig.~\ref{fig:subfig3}, we can custom generate captions based on tokens in nearly any position, which is a capability challenging for AR image captioning models.

\begin{figure*}[t!]
    \centering
\includegraphics[width=0.9\textwidth]{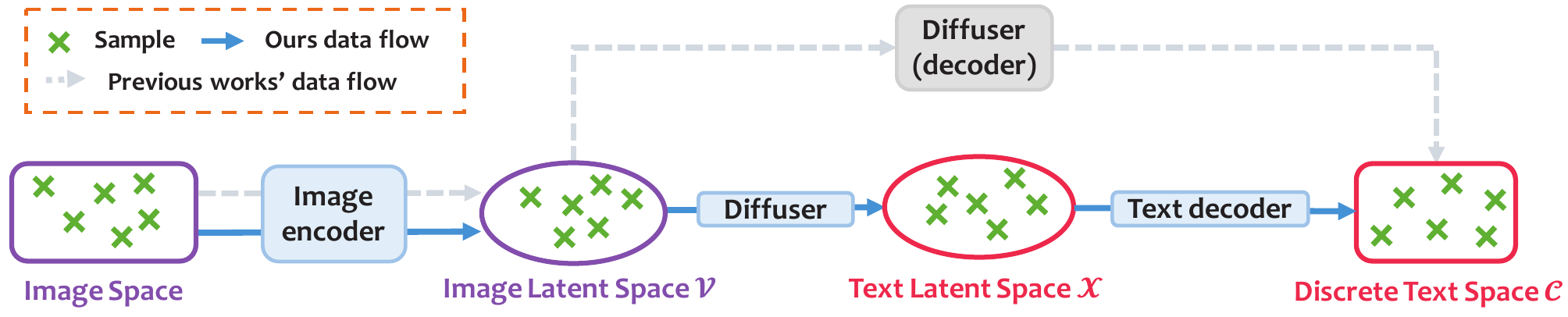}
    \caption{Comparison of the pipeline between our \modelname and that of previous diffusion-based models. We introduce text latent space to alleviate the burden on the diffuser. }
    \label{fig:high_level}
\end{figure*}

Despite the above benefits, the underperformance of previous diffusion models on image-to-text tasks hinders their popularity. 
Upon examining these diffusion-based models, we deduce that their unsatisfactory performance primarily stems from two factors: \textbf{(\uppercase\expandafter{\romannumeral1}) Semantic gaps in translating from images to texts}, which manifests in two dimensions: 1) the gap between visual information and textual representation, and 2) the gap between high-level text semantics and specific words. 
The previous paradigm attempts to simultaneously address these two gaps (illustrated by the grey dotted line in Fig.~\ref{fig:high_level}), but this proves overly challenging, resulting in poor performance. 
\textbf{(\uppercase\expandafter{\romannumeral2}) Incompatibility between continuous diffusion technology (image generation) and discrete inputs (text generation)}. Specifically, classical continuous diffusion models naturally align with the pixel space but struggle to transition directly to the discrete text space. Additionally, generated images have a fixed size, while caption lengths vary, presenting another challenge for diffusion models in determining the boundaries of generated captions. 

Given these considerations, we meticulously design a novel architecture \modelname, a  \textbf{La}tent \textbf{Di}ffusion-based \textbf{C}aptioner, for further amplifying the capability of diffusion models in image-to-text generation.
As depicted in Fig.~\ref{fig:intro}b and Fig.~\ref{fig:high_level} (blue line), rather than directly generating discrete text from image representation,  we treat the diffuser as an interface translating image information to high-level text representation (sentence latent). This approach alleviates the diffusion model's burden, enabling it to leverage its powerful generation capabilities in high-level semantic spaces~\citep{Ramesh2022HierarchicalTI}. After that, we utilize a Non-Auto-Regressive (NAR) text decoder to generate discrete tokens from latent space. 
To address problems like variable length of text, we propose a post-processing module including normalization and reassignment procedures. During inference, we introduce a Back$\&$Refine technique to provide more interaction between tokens, thus yielding better performance.

We conducted experiments mainly on the COCO dataset~\citep{Lin2014MicrosoftCC} to validate our model's capabilities. Remarkably, without pretraining or external modules, our model achieves 38.2 BLEU@4 and 126.2 CIDEr, significantly surpassing both diffusion-based methods and traditional NAR models. In addition to the unique advantages discussed earlier, our model also matches the performance of well-established pretrained AR models and outperforms BLIP in image paragraph captioning. These results underscore the potent generative ability and immense potential of diffusion models in image-to-text generation. We hope that our work offers a fresh perspective, fostering future research on diffusion models for image-to-text generation or even other text-centered multimodal generation tasks.





\section{ Related Works}
\subsection{Diffusion Models and their Applications} \label{sec: diff related}

Diffusion models have recently emerged as powerful generative models, with representative foundational architectures such as DDPM~\citep{DDPM} and DDIM~\citep{ddim}. These methods gradually transform samples into Gaussian noise and train a model to recover them, presenting a simple and stable learning objective for addressing issues like posterior and mode collapse that challenge prior models like VAE~\citep{Kingma2013AutoEncodingVB} and GAN~\citep{Goodfellow2014GenerativeAN}.

The impressive generative capabilities of diffusion models have led to their application across a spectrum of fields, including image ~\citep{Ramesh2022HierarchicalTI, dai2023emu}, audio ~\citep{Liu2023AudioLDMTG,ju2024naturalspeech}, video~\citep{Blattmann2023StableVD, Bai2024UniEditAU}, 3D~\citep{poole2022dreamfusion,lee2024dreamflow}, and human avatar ~\citep{he2023gaia,hu2023animateanyone}, among others. Yet, their application to text is still in its initial state. How to adapt discrete tokens into a diffusion model is an ongoing challenge. Existing approaches for tackling this problem generally fall into two categories: \textbf{(1) Discrete Text Diffusion Models}~\citep{austin2021structured, diffuser, He2022DiffusionBERTIG}, which mimic the diffusion process on the discrete space by directly corrupting text with \texttt{[MASK]} tokens. \textbf{(2) Continuous Text Diffusion Models}~\citep{difflm, diffseq_wu_kong_2022, Dieleman2022ContinuousDF, Yuan2022SeqDiffuSeqTD, Lin2022TextGW}, which use continuous embeddings to represent each token and then perform the classical diffusion process. While these approaches demonstrate the feasibility of applying diffusion models to text generation and show comparability with AR methods, they are limited to unimodal representations and may overlook high-level overall semantics to some extent.  \citet{lovelace2023latent} explore the concept of a text latent space. However, its diffusion model, designed for predicting BART's~\citep{Lewis2019BARTDS} hidden states, still relies on an AR generation mechanism, which suffers from its issues like low inference efficiency.

\begin{figure*}
    \centering
    \includegraphics[width=1\textwidth]{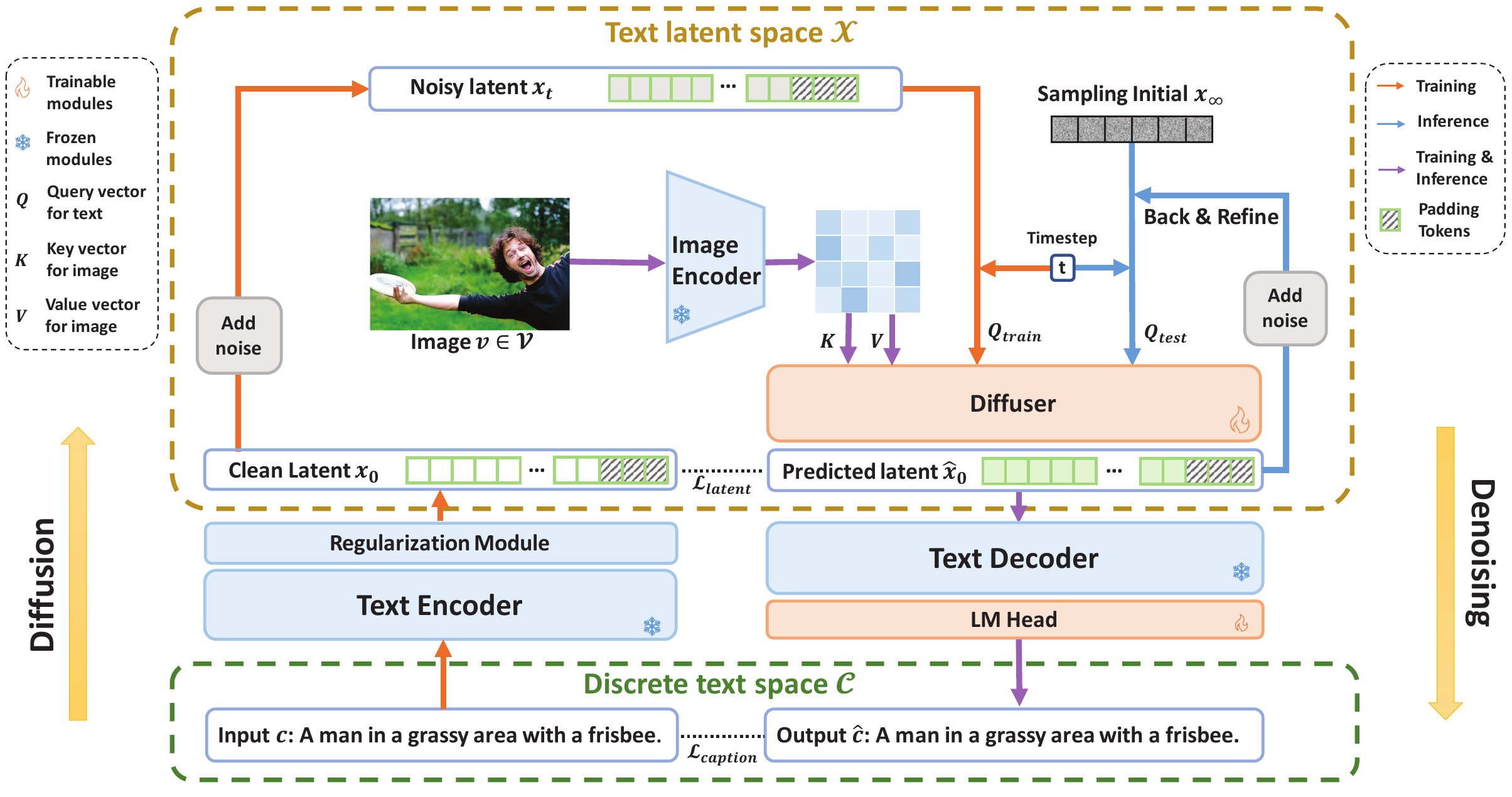}
    \caption{An overview of our \modelname model. It mainly consists of the Image Encoder, Text Encoder, Diffuser, and Text Decoder. The diffusion process is depicted on the left, while the denoising process is depicted on the right. Initially, the caption $c$ is encoded into a text latent $x_0$ by the text encoder. Subsequently, diffusion process occurs within the textual latent space $\mathcal{X}$, where a diffuser is trained to restore the noisy text latent $x_t$ to its clean counterparts $\hat{x}_0$, guided by the associated image. Finally, the denoised text latent $\hat{x}_0$ is passed through a NAR text decoder to generate the final caption $\hat{c}$.}
    \label{fig:main}
\end{figure*}
\subsection{Image-to-text Generation}
Image-to-text generation, especially the image captioning task, aims to describe the content of an image in natural language. Other task variants include dense captioning, which illustrates each object in the picture~\citep{densecap}, and paragraph captioning, which generates a detailed, lengthy paragraph~\citep{Krause2016AHA} and so on. 
Early AR approaches for captioning~\citep{Karpathy_Fei-Fei_2017, Vinyals2014ShowAT} employed an encoder-decoder architecture with a CNN (Convolutional Neural Network) to encode images and an RNN (Recurrent Neural Network) to generate captions. 
With the advent of Transformer~\citep{Vaswani2017AttentionIA} and large-scale pretraining methods, pretrained vision-language models~\citep{blip, vinvl, Li2020OscarOA, Ren2023TESTATT, ren2021iais, Zhao2023MMICLEV, Liu2023VisualIT} emerged and achieved high performance.

In contrast to the unidirectional generation of AR models, NAR models generate entire captions in parallel. MNIC~\citep{mnic} introduced the mask token strategy, and NAIC~\citep{NAIC} employed reinforcement learning in NAR caption generation. A special class of NAR methods, diffusion-based models has recently emerged. Most diffusion-based models~\citep{clipdifflm, He2023DiffCapEC, Liu2023PrefixdiffusionAL} follow the paradigm utilized in continuous diffusion models mentioned above. Additionally, Bit Diffusion~\citep{chen2022analog} encodes captions into binary bits, and DDCap~\citep{ddcap} applies a discrete diffusion model to captioning. SCD-Net~\citep{scd} is the state-of-the-art diffusion-based model with a semantic-conditional diffusion process. However, its cascaded architecture is relatively complex and requires an external retrieval module, limiting its further extension. Our work reexamines the diffusion-based paradigm and proposes a novel, compact architecture with improved performance.

\section{Methodology} \label{sec: 3method}

In this section, we introduce our diffusion-based image captioning model, \modelname. In \textsection~\ref{sec: overview}, we present the overall architecture of \modelname, including its training and inference pipeline. Subsequently, from \textsection~\ref{sec: method_latent} to \textsection~\ref{sec: back_refine}, we offer a detailed illustration.

\subsection{Overview} \label{sec: overview}

As illustrated in Fig.~\ref{fig:high_level}, we utilize a text encoder to transform the discrete text space $\mathcal{C}$ into a continuous text latent space $\mathcal{X}$. Subsequently, a diffuser is trained to serve as a bridge between the image representation space $\mathcal{V}$ and the text space $\mathcal{X}$, and finally, a text decoder maps the text latent codes back to the discrete text. 
\paragraph{Training Procedure.}
As shown in Fig.~\ref{fig:main}, given an image $v\in \mathcal{V}$ and its corresponding caption $c\in \mathcal{C}$, we encode the caption $c$ into the latent code $x_0 \in \mathcal{X}$. Subsequently, we train the diffusion model to fit the distribution of space $\mathcal{X}$. Initially, various levels of noise (introduced by different timesteps $t$) are added to $x_0$ to generate a noisy version $x_t$ (left panel). The diffuser then functions as a denoiser, aiming to recover $x_0$ conditioned on the image $v$ (right panel). This process can be represented by a function $f: x_t \xrightarrow{v} x_0$. 
\paragraph{Inference Procedure.}
During inference, $x_t (t\rightarrow \infty)$ is replaced with pure Gaussian noise $x_\infty \sim N(\mathbf{0},\mathbf{I})$. After that, $x_\infty$ is iteratively denoised by the diffuser $f$, resulting in $x_\infty \xrightarrow{v} \hat{x}_0$, where $\hat{x}_0$ represents the predicted text latent code. Finally, the decoder converts the predicted latent code back into discrete text $\hat{c} \in \mathcal{C}$.

\subsection{Latent Space Tailored for Text} \label{sec: method_latent}
As discussed in \textsection~\ref{sec: intro}, the text latent space $\mathcal{X}$ serves as a bridge between image space $\mathcal{V}$ and discrete text space $\mathcal{C}$, significantly alleviating the burden on diffusion models. Therefore, careful design of the $\mathcal{X}$ space is essential. This space should possess an appropriate semantic density to facilitate the semantic conversion from images to text.

Generally, the text latent space $\mathcal{X}$ is constructed by a text encoder. 
Depending on the selection of the text encoder, the previous work can be divided into two branches. 
The first branch~\citep{He2023DiffCapEC, Liu2023PrefixdiffusionAL} uses a very shallow encoder, e.g., a single embedding layer of BERT~\citep{Devlin2019BERTPO} to convert the discrete text into a continuous form. However, this method lacks interaction between tokens and overall semantic modeling, which poses a challenge when aligning images with these independent token embeddings. 
The second branch~\citep{Tang2023AnytoAnyGV, versatile} utilizes the entire BERT model as a text encoder, yielding densely packed semantic representations for sentences. However, compared to sentence latent with high information density, images typically possess much lower information density, characterized by substantial redundancy in pixel data~\citep{He2021MaskedAA, Ren2023TESTATT}. This discrepancy between the information densities of images and texts impedes the diffuser's ability to learn image-to-text translation effectively.

In contrast to these two methods, we opt to split the BERT model in half from the middle, utilizing the lower part as the text encoder and the upper part as the NAR text decoder. Setting the text latent space based on the middle layer of BERT yields improved alignment between vision and language, thereby enhancing performance. In addition, to improve the decoder's ability to reconstruct the text space, we make the parameters in the language model head trainable.
To achieve a more standardized sentence feature space conducive to noise addition, we employ normalization acting on this space. 
We gather a subset of all captions from the dataset and compute the mean and standard deviation of their corresponding latent codes, $\hat{\mu}(x)$ and $\hat{\sigma}(x)$. During training, these statistics are utilized to regularize the feature space of BERT by operating on each sample as follows: $
    \operatorname{norm}(x) 
    = [{x - \hat{\mu}(x)}]/[\hat{\sigma}(x) + \epsilon] 
$
.

We note that the lengths of target captions are variable, which necessitates the diffusion model to learn the end position of different captions during training. 
Considering the captions within a batch are padded to match the same sequence length, we can leverage the \texttt{[PAD]} token to determine the conclusion of a caption. 
This entails initially restoring the caption with \texttt{[PAD]} tokens and then eliminating all the \texttt{[PAD]} tokens to obtain the final caption.
However, our diffusion models operate in a text latent space $\mathcal{X}$, i.e., the feature space of the half BERT model, wherein the \texttt{[PAD]} token's feature is fused with contextual information, impeding its restoration to a \texttt{[PAD]} token by the text decoder. 
To address this issue, we extract all the special tokens like \texttt{[CLS]}, \texttt{[SEP]}, \texttt{[PAD]} in a caption, whose latent will be messy in the space $\mathcal{X}$, to form a set $\mathcal{S}$. We then reassign an empty latent $\boldsymbol{0}$ to the latent of these special tokens, as demonstrated in Eq.~\ref{equ: ass}. Here, $\tilde{x}^{(i)}$ represents the $i$-th position of the final regularized latent.
\begin{equation} \label{equ: ass}
\begin{aligned}
    \tilde{x}^{(i)}
    =\begin{cases}
    [\operatorname{norm}(x)]^{(i)} &\quad i \notin \mathcal{S}\\
     \boldsymbol{0},  &\quad i \in \mathcal{S}
    \end{cases}
\end{aligned}
\end{equation}
Through this technique, for short captions with pad tokens at the end, the diffuser can quickly identify this repeated pattern and easily recover these unified zero vectors, implicitly learning sentence boundaries. This approach avoids the need for an additional module for predicting sentence length~\citep{ddcap}. 
During inference, the token predicted as \texttt{[PAD]} can be easily erased by post-processing to generate various length captions. 

\subsection{Diffuser Mapping Image to Text} \label{sec: diffuser}
As shown in Fig.~\ref{fig:high_level}, the diffuser serves as an interface transforming the vision space $\mathcal{V}$ into the text latent space $\mathcal{X}$. To fit the distribution of space $\mathcal{X}$ by diffusion models, firstly we sample $x_t$, the noisy version of the latent code $x_0 \in \mathcal{X}$, as $x_t|x_0 \sim \mathcal{N}( \sqrt{\bar\alpha_t}x_0, \sqrt{1-\bar\alpha_t}\mathbf I)$, where $\bar{\alpha}_t=\prod_{i=1}^t \alpha_i = \prod_{i=1}^t (1-\beta_i)$ and $\beta_t \in (0,1)$ is the variance schedule. A notable property of this setting is that as $t\to \infty$, $x_t$ is equivalent to an isotropic Gaussian distribution, aligning with the starting state of inference. Then for the denoising process, we use diffuser to predict the original $x_0$ based on the image directly, denoted as $\hat{x}_0 = f_\phi(x_t, v, t)$, where $\phi$ represents the parameters of the diffuser. A rigorous mathematical explanation of the diffusion model can be found in App.~\ref{app:diff_math}.

\begin{figure}
    \centering
\includegraphics[width=1\linewidth]{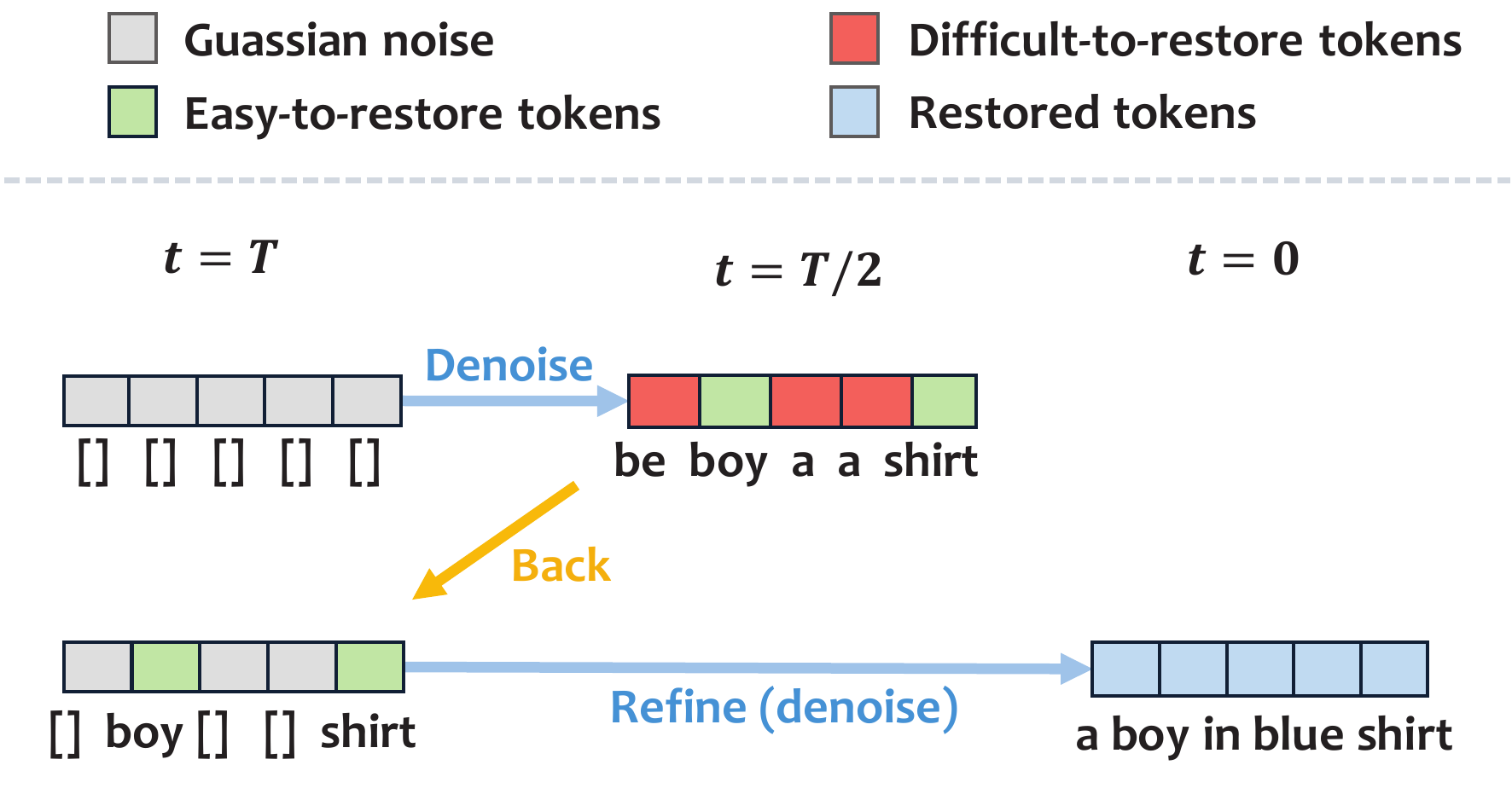}
    \caption{Illustration of Back$\&$Refine technique.}
    \label{fig:backrefine}
\end{figure}

To integrate visual information into the denoising process, we employ the cross-attention mechanism, where text serves as the query to extract information from related image patches. In contrast to previous approaches that typically inject image information by appending the \texttt{[CLS]} token of the vision encoder to text~\citep{clipdifflm, He2023DiffCapEC}, this approach achieves a better modality fusion effect, as will be verified in the ablation study. Additionally, we adapt classifier-free guidance~\citep{classifier_free} to this task to improve the alignment between text and image by randomly dropping some images during training. During inference, a linear combination of the conditional estimate $f_\phi(x_t, v, t)$ and unconditional one $f_\phi(x_t, \emptyset, t)$ is performed: $\hat{x}_0 = (1 + w) f_\phi(x_t, v, t) - w f_\phi(x_t, \emptyset, t)$  where $w$ is a hyper-parameter.

We use a two-fold loss to train the diffuser in \modelname. The first one is the loss $\mathcal{L}_{\text{latent}}$, operating within the text latent space. This loss calculates the Mean Squared Error (MSE) between the predicted text latent $\hat{x}_0= f_\phi(x_t, v, t)$ and the original text latent $x_0$. 
The second one is the cross-entropy loss $\mathcal{L}_{\text{caption}}$ in the discrete caption space. 
Specifically, let $\hat{x}_0^{i}$ be the $i$-th position of the text latent and $w^i$ be the $i$-th word in the ground-truth caption. 
Given $\hat{x}_0^{i}$, the probability of correctly predicting $w^i$ in the vocabulary is $p_\theta(w^{i} | \hat{x}_0^{i} )$, where $\theta$ represents the parameters of the diffuser and language model head in text decoder. 
This loss makes the output of the caption diffuser shrink faster, sharing the same intuition with XE loss in~\citep{scd} and anchor loss in~\citep{Gao2022DifformerED}. Meanwhile, it also helps train the language model head in the decoder. 
In summary, the final loss $\mathcal{L}$ is: 
\begin{equation} \label{equ: loss}
\begin{split}
    \mathcal{L} &= \mathcal{L}_{\text{latent}} +  \lambda \mathcal{L}_{\text{caption}} \\
    &= \| f_\phi(x_t, v, t) - x_0\| - \lambda \prod_{i=1}^{n} p_\theta(w^{i} | \hat{x}_0^{i} ),
\end{split}
\end{equation}
where $\lambda$ is a hyper-parameter.

\subsection{Back$\&$Refine Technique during Inference}\label{sec: back_refine}

\begin{table*}[] \small
\centering
\begin{adjustbox}{max width=\textwidth}
\begin{tabular}{@{}lr|ccccccc@{}} \midrule
                  Model  & \# Images  & BLEU@4  & CIDEr     & METEOR    & SPICE    & ROUGE-L    & CLIP-Score & BERT-Score \\ \midrule
\rowcolor{dt!50}
\multicolumn{9}{l}{\textit{Autoregressive}}                                                                                          \\ \midrule
Show and Tell~\citep{Vinyals2014ShowAT}       & -                              & 31.4 & 97.2  & 25.0 & 18.1 & 53.1 &  69.7  &    93.4                 \\
CLIPCap~\citep{Mokady2021ClipCapCP}                       &            -        & 33.5 & 113.1 & 27.5 & 21.1 & -    &    -        &  -          \\
\grayrow{OSCAR$\dag$~\cite{Li2020OscarOA}} & \grayrow{7M} & \grayrow{36.5} & \grayrow{123.7} & \grayrow{30.3} & \grayrow{23.1} & \grayrow{-}   & \grayrow{-}        & \grayrow{-}        \\
\grayrow{ViTCap$\dag$~\cite{vitcap}}        & \grayrow{4M} & \grayrow{36.3} & \grayrow{125.2} & \grayrow{29.3} & \grayrow{22.6} & \grayrow{58.1} & \grayrow{-}        & \grayrow{-}        \\
\grayrow{VinVL$\dag$~\cite{vinvl}}          & \grayrow{6M} & \grayrow{38.2} & \grayrow{129.3} & \grayrow{30.3} & \grayrow{23.6} & \grayrow{60.9} & \grayrow{76.6}     & \grayrow{88.5}     \\
\grayrow{BLIP$\dag$~\cite{blip}}            & \grayrow{129M} & \grayrow{39.7} & \grayrow{133.3} & \grayrow{-}   & \grayrow{-}    & \grayrow{-}    & \grayrow{77.4}     & \grayrow{94.4}     \\
\grayrow{GIT$\dag$~\cite{Wang2022GITAG}}    & \grayrow{4M} & \grayrow{40.4} & \grayrow{131.4} & \grayrow{30.0} & \grayrow{23.0} & \grayrow{-}    & \grayrow{-}        & \grayrow{-}        \\
\midrule
\rowcolor{dt!50}
\multicolumn{9}{l}{\textit{Traditional Non-autoregressive}} \\
NAIC$\rm{_{KD}}$~\citep{NAIC}              & 0.1M                         & 28.5 & 98.2  & 23.6 & 18.5 & 52.3 &    -        &  -          \\
MNIC~\citep{mnic}                & 0.1M                        & 31.5 & 108.5 & \textbf{27.5} & \textbf{21.1} & \textbf{55.6} &  -          &  -          \\

FNIC~\citep{fnic}                & 0.1M                            & \textbf{36.2} & \textbf{115.7} & 27.1 & 20.2 & 55.3 &  -          &  -          \\ \midrule
\rowcolor{dt!50}
\multicolumn{9}{l}{\textit{Diffusion model based}}                                                                                   \\ \midrule
DiffCap~\citep{He2023DiffCapEC}             &   0.1M                       & 31.6 & 104.3 & 26.5 & 19.6 & 55.1 &  73.6*          &   92.2*         \\
Bit Diffusion~\citep{Chen2022AnalogBG}       & 0.1M                           & 34.7 & 115.0 & -    & -    & 58.0 &    -        &      -      \\

DDCap~\citep{ddcap} &      0.1M                & 35.0 & 117.8 & 28.2 & 21.7 & 57.4 &  74.1*          &     93.4*       \\

SCD Net~\citep{scd}             & 0.1M                            & 37.3 & 118.0 & 28.1 & 21.6 & 58.0 &     74.5*       &      93.4*      \\
\textbf{\modelname} (ours, \textbf{5 steps})            &       0.1M                     & 35.1  & 115.2   &  27.4     &  21.3    & 56.7     &   77.1   &    93.8          \\ 
\textbf{\modelname} (ours, \textbf{30 steps})            &      0.1M                      & \textbf{38.2}  & \textbf{126.2}  &  \textbf{29.5}     &  \textbf{22.4}    & \textbf{58.7}   &   \textbf{77.3}   &    \textbf{94.4}          \\ 
\bottomrule     
\end{tabular}
\end{adjustbox}
\caption{Model performance on COCO dataset. 
$\dag$ indicates pretrained models, and we gray them out since they use much more training data. * represents the results of models reproduced by ourselves. 
 Our model achieves state-of-the-art performance across various metrics for both diffusion-based and traditional NAR models, and exhibits comparable performance with some well-established pretraining auto-regressive frameworks, despite being trained on significantly less data. 
 The inference time measured on an A100 GPU for 5 steps is 0.020 s$/$img and 30 steps is 0.105 s$/$img.
 }
\label{tab: main}
\end{table*}
\citet{chen2022analog} noted that the traditional inference process of diffusion models often discards the previously estimated value $\hat{x}_0$ at each subsequent time step, which leads to suboptimal results.  
We notice that this situation occurs not only in the temporal dimension (time step) but also in spatial dimensions, i.e., the positions of words within a sentence.  In contrast to AR models, which explicit sequential dependencies across tokens, the diffusion model emits all tokens in parallel. 
Considering that some tokens, such as the main objects in the image, can be easily restored. Conversely, some tokens, representing visual details, are challenging to restore. Adding the same scale of noise to the easy-to-restored tokens as the others is unreasonable and inefficient. Hence, it is intuitive to leverage these easily restored informative tokens as conditions to aid in the refinement process of the more challenging ones. 

Accordingly, we propose a method named Back$\&$Refine to rollback and refine those challenging tokens with low prediction confidence. As illustrated in Fig.~\ref{fig:backrefine}, given a sentence with a sequence length $L$ and a sampling step $T$ to be restored. At time $T/2$, we calculate the model's prediction confidence for all tokens. We keep $L/2$ tokens with the highest confidence (depicted in green), as they are considered easily restorable and relatively accurate. 
Conversely, for the remaining $L/2$ tokens with the lowest confidence (in red), we reproduce them by noising them with complete Gaussian noise (in grey). Then we set the current $t = T$ and start a new denoising procedure to restore the challenging tokens based on the easier ones. Applying this technology, we observe performance gains, and
we also provide another example for deeper understanding Back$\&$Refine in App.~\ref{sec: app_correct}.


\section{Experiments} \label{sec: exp}
\subsection{Experimental Settings}
\paragraph{Dataset and Metrics}

We conduct our experiments on MS COCO Karpathy split~\citep{Lin2014MicrosoftCC, Karpathy2014DeepVA}, which comprises 113,287 training images, 5,000 validation images, and 5,000 test images. Each image is associated with 5 reference captions. For evaluation, we follow the common practice and use several metrics including BLEU@4~\citep{papineni-etal-2002-bleu}, CIDEr-D~\citep{Vedantam2014CIDErCI}, METEOR~\citep{Banerjee2005METEORAA}, ROUGE-L~\citep{Lin2004ROUGEAP}, and SPICE~\citep{Anderson2016SPICESP}. Additionally, we employ two model-based metrics: CLIP-Score~\citep{Hessel2021CLIPScoreAR} to assess semantic alignment between generated captions and images, and BERT-Score~\citep{bertscore} to evaluate text quality.

\paragraph{Implementation Details}

In our \modelname model, the encoder and decoder are frozen, except for the LM-head. The weights of the encoder and decoder are initialized from the bottom 6 layers and top 6 layers of $\rm{BERT_{base}}$, respectively. The rationale for selecting such a latent space is explained in App.~\ref{sec: app_layers}. For the diffusion forward process, we employ the widely used cosine $\beta$ schedule and adopt the noise factor~\citep{Gao2022DifformerED}. The diffuser consists of 12 transformer encoder blocks with additional cross-attention layers in each block, and the weights are randomly initialized. To extract image features, we use the pretrained image encoder from $\rm{BLIP_{base}}$~\citep{blip}, which employs ViT-B/16, for a fair comparison with BLIP. The model is trained on 8$\times$V100 GPUs for 60 epochs with a peak learning rate of 5e-5 and a warmup ratio of 0.1. More details can be found in App.~\ref{sec: app_hyper}.

\subsection{Quantitative Analysis}

We benchmark our \modelname model against prior baselines, encompassing auto-regressive, traditional non-autoregressive, and diffusion-based models, leveraging the COCO dataset. As shown in Tab.~\ref{tab: main}), our model achieves state-of-the-art performance across various metrics for both diffusion-based and traditional NAR models. Specifically, \modelname achieves 38.2 BLEU@4 and 126.2 CIDEr, marking improvements of 0.9 and 8.2, respectively, compared to the previous SOTA method, SCD-Net. Remarkably, a variant of our model, utilizing only 5 inference steps, even outperforms all prior diffusion-based models in both CLIP-Score and BERT-Score. Moreover, it is noteworthy that \modelname exhibits comparable performance with well-established pretraining auto-regressive frameworks such as ViTCap and VinVL, despite being trained on significantly less data.

To evaluate our model's capacity for considering holistic context, we tackle the task of image paragraph captioning~\cite{Krause2016AHA} to generate a multi-sentence description of an image. Our model seamlessly adapts to paragraph captioning by extending the predefined length without additional special designs. Training our model on the dataset from~\cite{Krause2016AHA} yields a BLEU@4 score of 7.3, surpassing finetuned BLIP's 6.1 and highlighting our model's advantage in mitigating error accumulation (refer to App.~\ref{sec: app_para} for more details). All these quantitative indicators above substantiate the accuracy and high quality of the captions generated by our model.

\subsection{Case Study and Human Evaluation}

\begin{figure}
    \centering
    \includegraphics[width=0.9\linewidth]{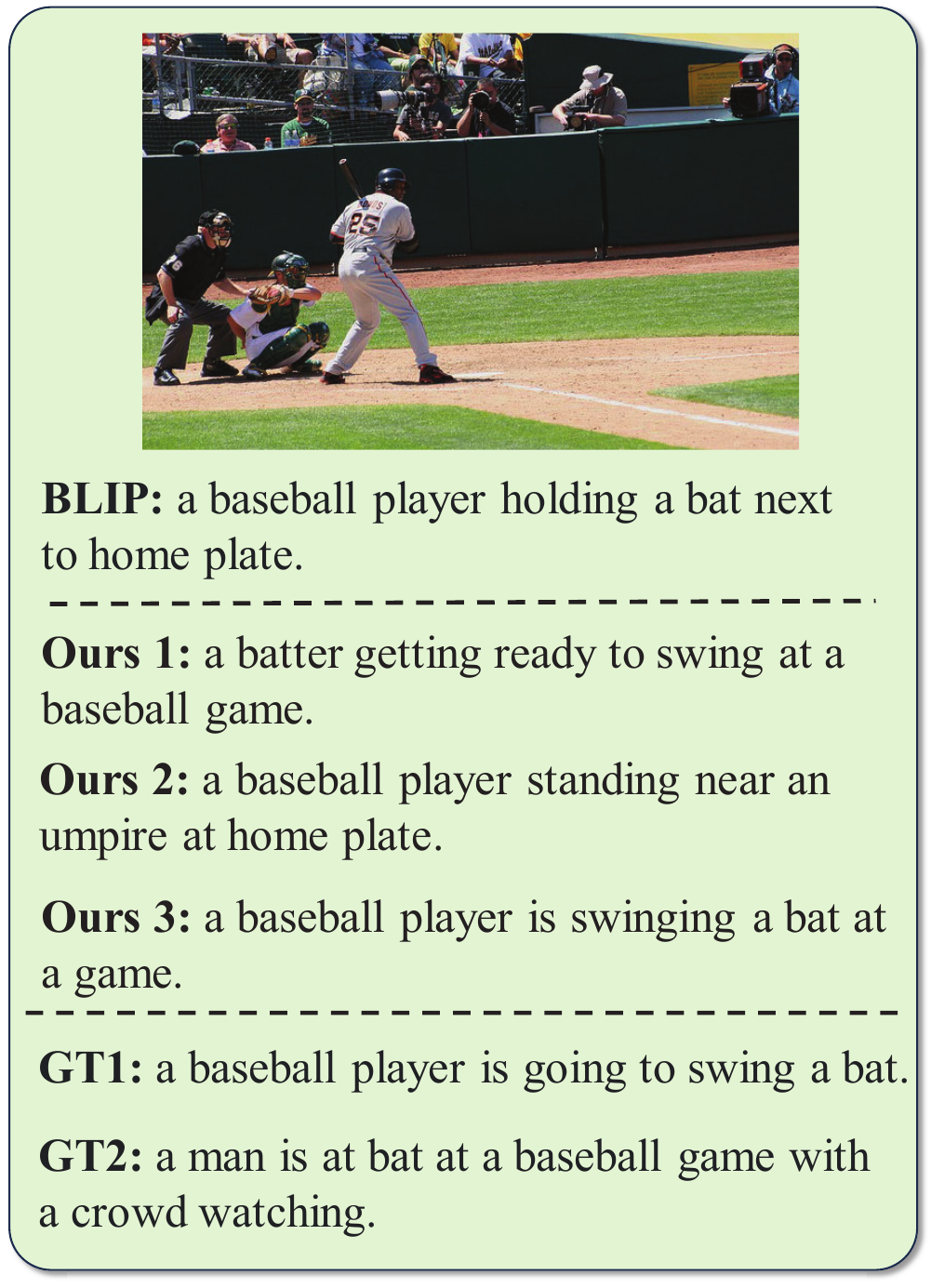}
    \caption{An example generated by our model.}
    \label{fig:case}
\end{figure}

\begin{figure}
    \centering
    \includegraphics[width=0.9\linewidth]{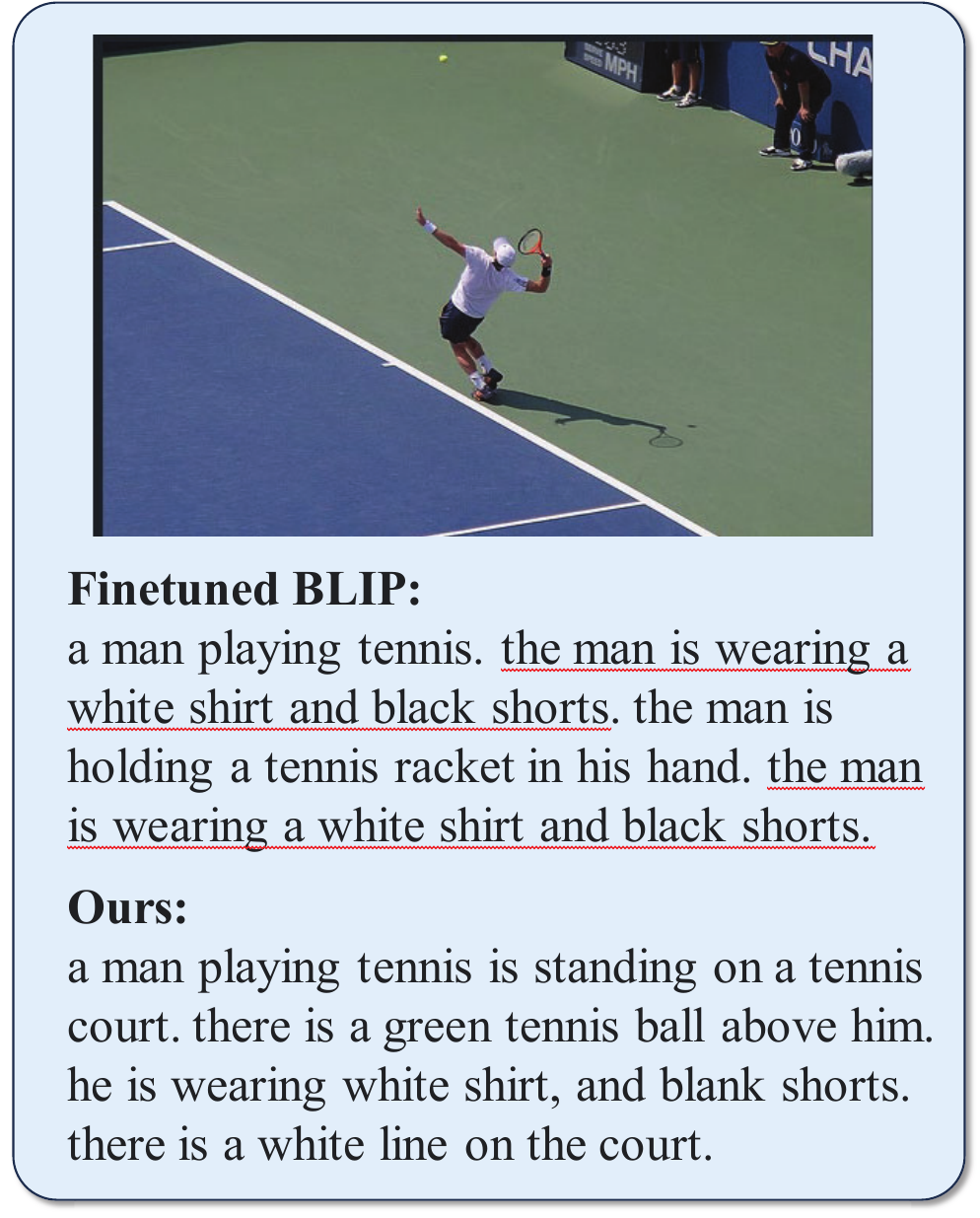}
    \caption{An example generated by fine-tuned BLIP model and ours in image paragraph captioning.}
    \label{fig:para_contrast}
\end{figure}

We conduct a case study to illustrate the faithfulness and diversity of the captions generated by \modelname. As depicted in Fig.~\ref{fig:case}, the generated captions are not only reasonable and fluent but also exhibit inherent diversity due to the varied sampling noises introduced at the start of inference. Additional examples can be found in App.~\ref{sec: app_coco_res}. In the image paragraph captioning task, Fig.~\ref{fig:para_contrast} highlights a notable disparity between the two approaches. 
While BLIP-generated captions exhibit good quality, the sentences in the output often seem disjointed, with occasional repetitions and a tendency to start with `the man'. Conversely, by leveraging a broader context, our model produces sentences with a more cohesive logical relationship.

We conduct user studies to evaluate the generated captions of \modelname, inviting volunteers to rate captions on a five-point scale (1-5) for accuracy, conciseness and fluency. The results, presented in Tab.~\ref{tab: human_eval}, demonstrate that our model surpasses the previous diffusion-based state-of-the-art SCD-Net in both aspects and achieves comparable results with BLIP. Details can be found in App.~\ref{sec: human_eval}.
\begin{table} \small
\centering
\begin{adjustbox}{max width=0.9\linewidth}
\begin{tabular}{c|ccc}
\toprule
Model           & SCD-Net         & BLIP  & Ours \\ \midrule
Fluency & 2.8 &     \textbf{4.9}  &     4.5       \\ 
\midrule
Accuracy & 3.3 &    4.2   &    \textbf{4.4}        \\
\midrule
Conciseness & 3.4  &  4.4     &    \textbf{4.7}        \\
\bottomrule
\end{tabular}
\end{adjustbox}
\caption{Results of human evaluation. }
\label{tab: human_eval}
\end{table}

\subsection{Unleashing the Speed of Diffusion Model}

Despite their powerful generative capabilities, diffusion models are notorious for their slow inference speed. Most previous works require more than 50 inference steps, significantly slower than traditional NAR methods, which typically involve around 10 refinement procedures. However, as shown in Tab.~\ref{tab: main}, our model achieves remarkable performance even with just 5 steps. We attribute this surprising convergence speed to specific techniques employed in our \modelname model. Firstly, the direct prediction of $x_0$ and the definition of caption loss enable the model to rapidly learn the distribution of discrete caption text, akin to the consistency model~\citep{Song2023ConsistencyM}. Secondly, the carefully selected noise schedule and noise factor significantly enhance the learning process of diffusion models. Regarding observed latency, the results in Tab.~\ref{tab: time} (measured on a single A40 GPU with a batch size of 256) and Fig.~\ref{fig:subfig1} demonstrate that our model showcases a rapid inference speed, excelling not only in the domain of diffusion-based models but also when compared to auto-regressive models.


\subsection{Customizing the Generation Process} \label{sec: custom}

In contrast to the unidirectional generation manner of AR models, our \modelname model adeptly fills in empty words at almost any position within a sentence, harnessing its capability to capture more holistic information, as demonstrated in Fig.~\ref{fig:subfig3}. Technically, when provided with a caption containing blanks, we extract contextual embeddings of the given tokens and mask the blank tokens with Gaussian noise. The standard denoising process is then applied, with the exception of reinserting the embeddings of predefined tokens back to their respective positions after each inference step, ensuring that the given information is retained. Through this method, our model functions as a customized generator based on the provided tokens. Additional results can be found in App.~\ref{sec: app_custom}.
\begin{table}\small
\centering
\begin{adjustbox}{max width=\linewidth}
\begin{tabular}{c|cccc}
\toprule
Model           & DiffCap         & DDCap  & Ours \\ \midrule
Inference latency(s/img) & 0.625 &  0.113     &  \textbf{0.049}          \\ \bottomrule
\end{tabular}
\end{adjustbox}
\caption{Inference latency of diffusion-based models. }
\label{tab: time}
\end{table}
 \begin{table} 
\centering
\begin{adjustbox}{max width=\linewidth}
\begin{tabular}{c|ccc}
\toprule
    & BLEU-4 & CIDER \\ 
    \midrule
w/o Back $\&$ Refine              & 37.3   & 121.5  \\ \midrule
$t=0.5T$; $l=0.5L$ (Our final version) & \textbf{38.2}   & \underline{126.2}  \\ \midrule
$t=0.8T$; $l=0.5L$              & \underline{38.1}   & \textbf{126.6}  \\ \midrule
$t=0.2T$; $l=0.5L$              & 37.5   & 122.9  \\ \midrule
$t=0.5T$; $l=0.8L$              & 37.6   & 123.5  \\ \midrule
$t=0.5T$; $l=0.2L$              & 37.5   & 122.3  \\
\bottomrule
\end{tabular}
\end{adjustbox}
\caption{Results of different settings of Back$\&$Refine Technique. The second best result is underlined.}
\label{tab:back_refine}
\end{table}
\subsection{Analysis for the Back$\&$Refine Technique}
Regarding the Back$\&$Refine technique, as discussed in \textsection~\ref{sec: back_refine}, we specify that when predicting a sentence with a sequence length of $L$ and a sampling step of $T$, we opt to backtrack at time $t=0.5T$ ($T\to0$) and discard $l=0.5L$ tokens. 
We present an experiment to investigate the influence of the two ratios in Back$\&$Refine (applied once). As shown in Tab.~\ref{tab:back_refine}, choosing an early backtracking time ($t=0.8T$) leads to inadequate recovery of easy-to-restore tokens, failing to provide sufficient information and resulting in performance just similar to scenarios without Back$\&$Refine. Conversely, backtracking at a late time ($t=0.2T$) does not yield significant improvement, as easy-to-restore tokens are typically recovered quickly, and the additional steps introduce an undesirable drop in inference speed. Similarly, dropping the majority of tokens in the Back procedure ($0.8L$) results in a situation akin to scenarios without Back$\&$Refine. Dropping too few tokens ($0.2L$) may introduce many mistaken tokens, adversely affecting performance. Therefore, we deliberately choose the setting $t=0.5T$ and $l=0.5L$ for our final version.
Furthermore, an alternative option is to establish a confidence score threshold instead of directly dropping the last half. However, in the majority of our early experiments, these two settings exhibit a negligible performance gap. Consequently, we opt for the simpler second method in our final version.

\begin{table}[]  
  \small
  \begin{adjustbox}{max width=\linewidth}
  \centering
  \begin{tabular}{@{}c cccccc|c|c@{}}
    \toprule   
    \#Row  & \begin{tabular}[c]{@{}c@{}}Cross\\ attn.\end{tabular}  & \begin{tabular}[c]{@{}c@{}}Cap.\\ loss\end{tabular} & PLM  & \begin{tabular}[c]{@{}c@{}}Norm\\ Reass\end{tabular} & Split  & B\&R & B@4  &C  \\    
    \midrule
    a  &&&&&&& 15.4 & 46.3  \\ 
    b  &$\checkmark$&&&&&& 20.3  & 59.1   \\ 
    c  &$\checkmark$&$\checkmark$&&&&& 22.8 & 76.3  \\ 
    d  &$\checkmark$ &$\checkmark$&$\checkmark$&&&& 26.9 & 91.8 \\

e&$\checkmark$&$\checkmark$&$\checkmark$&$\checkmark$&&&31.6&103.5 \\
    f&$\checkmark$&$\checkmark$&$\checkmark$&$\checkmark$&$\checkmark$&&33.4&110.0 \\
    g&$\checkmark$&$\checkmark$&$\checkmark$&$\checkmark$&$\checkmark$&$\checkmark$& \textbf{34.1} &\textbf{113.4}\\
    \bottomrule
  \end{tabular}
 
  \end{adjustbox}
  \caption{Ablation on COCO dataset.}
  \label{tab: component}
\end{table}

\subsection{Ablation study}

To validate the effectiveness of our core designs, we conduct ablation studies on the COCO dataset.  Owing to the extensive time required for the ablation study, we opted for a subset of the dataset and trained all models for 40 epochs, (at which point the validation loss has already converged). For the inference phase, we performed 5 steps.  We begin with a simple baseline that appends only the \texttt{[CLS]} token of the image feature to the end of text embeddings and then trains the diffuser to recover them. Subsequently, we progressively incorporate our proposed techniques to evaluate their performance. As depicted in Tab.~\ref{tab: component}, all modules exhibit performance gains. The use of PLM (BERT) and regularization in this space significantly enhances performance, emphasizing the importance of a refined latent space. Techniques aimed at better capturing visual information, such as cross-attention and splitting the BERT, also play pivotal roles in improving performance.

\section{Conclusion}

In this paper, we reexamine the diffusion-based image-to-text paradigm and introduce a novel architecture, denoted as \modelname. Our model attains state-of-the-art performance among diffusion-based methods and demonstrates comparable capabilities with some pre-trained AR models. Moreover, our extensive experiments reveal the exciting advantages of diffusion models over AR models in considering more holistic contexts and emitting all tokens in parallel. Consequently, we posit that diffusion models hold substantial potential for image-to-text generation and we hope that our work will open new possibilities in this field.

\section*{Limitations}

For simplicity and focus, this paper concentrates on the main research topic of image-to-text generation. Nevertheless, we observe that our model can be readily adapted to other modalities or even pure text generation with minimal modifications. We leave these potential extensions for future work, and meanwhile, we hope this paper will inspire confidence among researchers engaging in text-centered multimodal generation tasks with diffusion models and look forward to exciting future works in this area. Furthermore, due to resource constraints, the model parameters and datasets employed in our study are not extensive. Considering the remarkable emergent abilities demonstrated by scaling up autoregressive models like GPT, it becomes an intriguing and worthwhile exploration to investigate whether our model or general diffusion models, can exhibit similar scalability.

\paragraph{Risk Consideration:} As a generative model, our model may inadvertently produce results that are challenging to distinguish from human-written content, raising concerns about potential misuse. Employing text watermark techniques could be beneficial in mitigating this issue. Additionally, diffusion models typically demand substantial computational resources for training, leading to increased carbon dioxide emissions and environmental impact.

\bibliography{anthology, custom}

\newpage
\appendix
\section{Additional Results}
\label{sec:reference_examples}

\subsection{Generated Samples from COCO Dataset} \label{sec: app_coco_res}
\begin{figure*}[ht]
    \centering
    \includegraphics[width=1\linewidth]{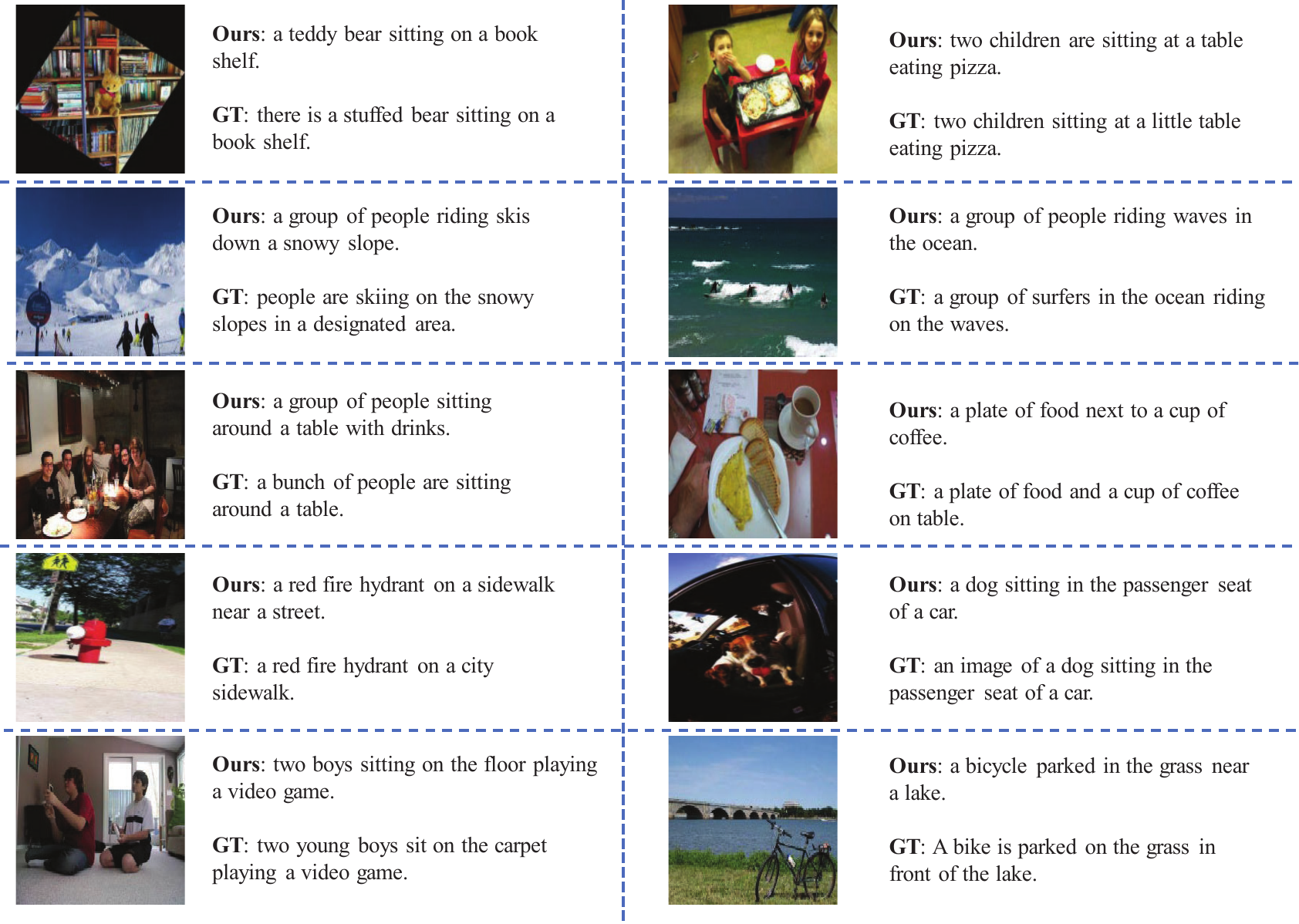}
    \caption{More examples generated by our model on COCO datasets.}
    \label{fig:app_coco}
\end{figure*}

Additional examples generated by our \modelname model are presented in Fig.~\ref{fig:app_coco}. It is shown that our model adeptly captures the main objects and their relationships in the depicted images. Simultaneously, the generated captions exhibit a high level of fluency.

\subsection{Custom Generation} \label{sec: app_custom}

\begin{figure*}[ht]
    \centering
    \includegraphics[width=1\linewidth]{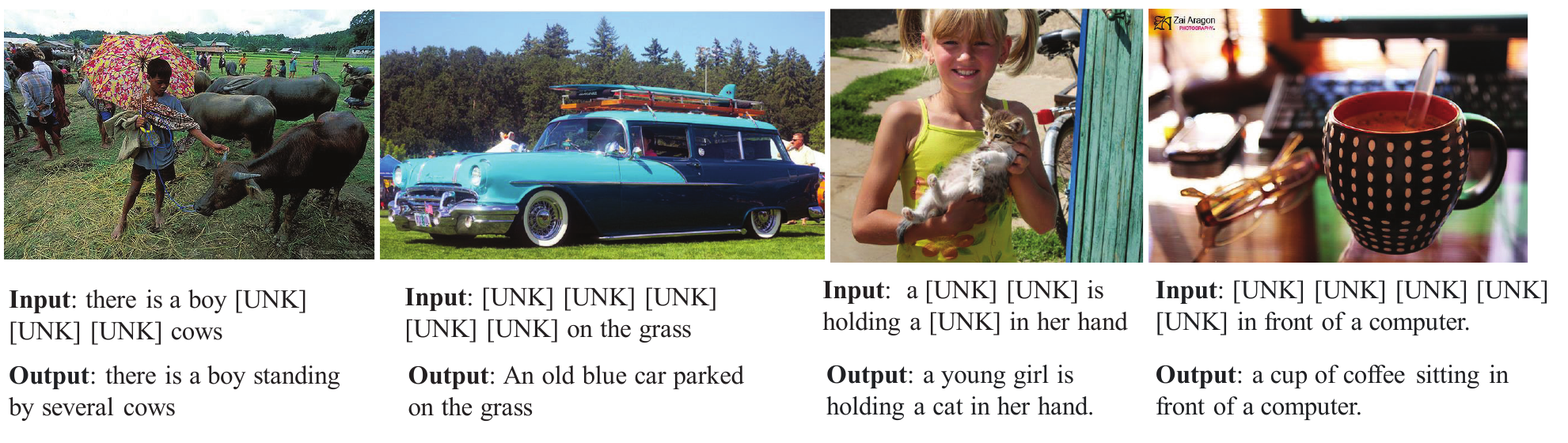}
    \caption{More examples of custom generation.}
    \label{fig:app_cust}
\end{figure*}

Utilizing the partially adding noise technique described in \textsection ~\ref{sec: custom}, we observed that, unlike the unidirectional generation approach of AR models, our \modelname model can effectively insert words into almost any position within a sentence. Fig.~\ref{fig:app_cust} offers additional examples to illustrate the generalization ability of this method.

\subsection{Gradual Denoising Process during Inference}
We show the gradual denoising process of our model. As a generative model,  the diffusion model is capable of modeling the distribution of any space by being trained to progressively transform random noise into a ground truth sample. In our model, we opt to apply diffusion to the latent space of text, i.e., the sentence feature space of BERT. As illustrated in \textsection~\ref{sec: intro}, our process begins with Gaussian noise. At each step, we subtract a certain amount of noise. As the number of steps increases, the denoised sentence feature converges towards the ground-truth sentence feature. Mathematically, our diffuser will model such a distribution:
$P(T_{i+1}|T_{i}, I)$. Here, $I$ represents image features, $T_i$ denotes the sentence feature at the last step $i$, and $T_{i+1}$ signifies the new sentence feature with reduced noise.

\begin{figure*}
    \centering
    \includegraphics[width=0.8\textwidth]{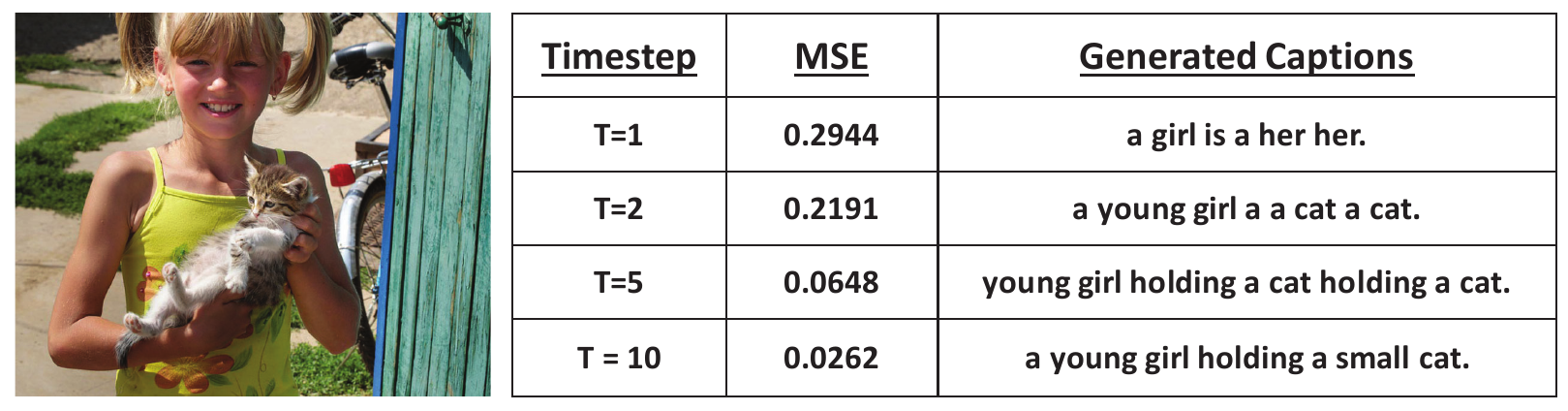}
    \caption{Gradual denoising process of diffusion models.}
    \label{fig:app_gradual}
\end{figure*}

As an illustrative example, refer to a specific case in Fig.~\ref{fig:app_gradual}. With an increase in steps, the Mean Squared Error (MSE) distance between the current sentence vector and the ground-truth sentence vector diminishes, and the sentences generated by the predicted sentence vector at each step become progressively more fluent. These findings collectively demonstrate the capability of our diffusion model to gradually steer noised sentence vectors towards ground-truth sentence vectors and generate high-quality samples. When we carefully check the generated captions, notably, the main objects initially emerge, and subsequently, more details are incrementally added, resulting in increasingly fluent sentences. This characteristic also serves as inspiration for our Back$\&$Refine Technique, as discussed in \textsection~\ref{sec: back_refine}.

\begin{figure}
    \centering
    \includegraphics[width=0.45\textwidth]{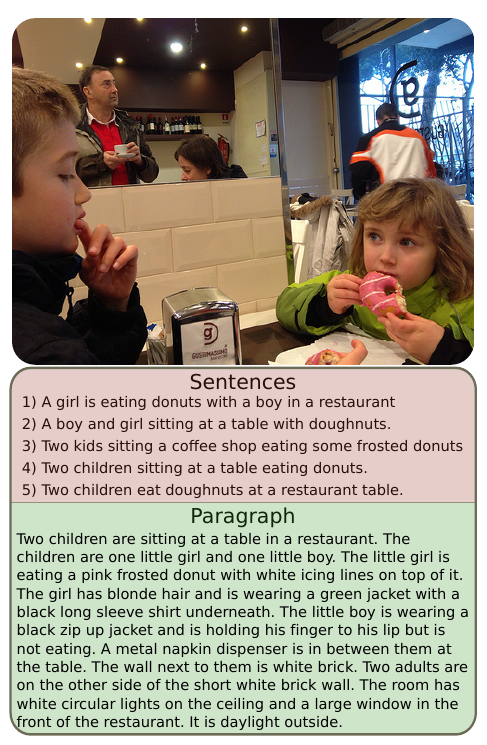}
    \caption{An example from image paragraph captioning dataset (cited from~\cite{He2021MaskedAA}).}
    \label{fig:app_para_case}
\end{figure}

\subsection{Exploration on the Choice of Latent Space} \label{sec: app_layers}

In \textsection~\ref{sec: method_latent}, we addressed the information density gap between vision and language by diffusing on the middle layer of the BERT model. Regarding the choice of different possible latent spaces, we conducted preliminary experiments to investigate this issue. We implemented various latent spaces extracted from different layers (specifically, the 3rd, 6th, and 9th layers of the $\rm{BERT_{base}}$), with the findings presented in Tab.~\ref{tab: bert_layers}. It is important to note that in these initial tests, the model was trained for 40 epochs without incorporating caption loss and the Back$\&$Refine technique. Our results indicate that the 6th layer outperforms the others, which is the rationale behind its selection as our final setting in the paper. Although we did not explore every layer, our preliminary experiments already provided us with a degree of confidence and suggested that layers proximal to the midpoint of BERT (while not necessarily exactly the 6th layer due to different datasets or hyperparameters) may have better alignment with image space.

\begin{table} 
\centering
\begin{adjustbox}{max width=0.7\linewidth}
\begin{tabular}{c|ccc}
\toprule
    & BLEU-4 & CIDER \\ 
\midrule
Layer 3 of $\rm{BERT_{base}}$         & 31.1   & 98.4         \\
\midrule
Layer 6 of $\rm{BERT_{base}}$         & \textbf{33.7}   & \textbf{112.3}        \\
\midrule
Layer 9 of $\rm{BERT_{base}}$         & 32.3   & 106.8 \\
\bottomrule
\end{tabular}
\end{adjustbox}
\caption{Performance for different layers of BERT. }
\label{tab: bert_layers}
\end{table}

\subsection{The Self-Correction Ability of Back$\&$Refine Technique} \label{sec: app_correct}
In our Back$\&$Refine technique, we utilize the preserved easy-to-restore tokens to facilitate the generation of hard-to-restore tokens. However, it is important to emphasize that the remaining tokens in the Back procedure still have opportunities for revision in the Refine procedure rather than being fixed. On the one hand, our initial experiments find that well-restored tokens are inclined to be preserved during the Refine procedure. This observation guides our intuition to leverage these well-restored tokens to enhance the denoising process for challenging-to-restore tokens. On the other hand, it's noteworthy that, as these well-restored tokens are also required to pass through the Refine procedure, there are still chances to address errors inadvertently retained, such as grammar issues. For example, in a real-case scenario, when the Back procedure finishes, a sentence is ``[] children is [] [] [] [] a pizza.'' where the ungrammatical word ``is'' is preserved. However, through the Refine procedure, the final output caption is corrected to ``Two children are sitting at a table eating pizza.''

\section{Additional Details in Experiments}
 
\subsection{Details about Experiments on Image Paragraph Captioning}  \label{sec: app_para}
The objective of image paragraph captioning is to generate comprehensive paragraphs that describe images, providing detailed and cohesive narratives. This concept was initially introduced in~\citep{Krause2016AHA}, where the authors proposed a dataset comprising 19,551 images from MS COCO~\citep{Lin2014MicrosoftCC} and Visual Genome~\citep{krishna2016visual}, each annotated with a paragraph description. An illustrative example is presented in Fig.~\ref{fig:app_para_case} (cited from~\cite{Krause2016AHA}).

To assess our model's ability to consider holistic context, we compare the performance of our model and BLIP on this task. For our model, we extend the predefined length to 60 and conduct training over 120 epochs. For BLIP, we fine-tune from BLIP$\rm{_{base}}$ using the same number of epochs and an initial learning rate of 1e-5. Subsequently, we evaluate the results using BLEU on the test set. In the case of BLIP, the maximum length is set to 60, and the number of beams is 5 during inference.

\subsection{Human Evaluation} \label{sec: human_eval}
As a generative task, in addition to automatic metrics, it is imperative to assess results through human subjective evaluation. To this end, we utilize MOS (Mean Opinion Score) as our metric and enlist the feedback of 20 experienced volunteers, who were tasked with rating results on a scale of 1-5. They evaluated the results from three perspectives: fluency, accuracy, and conciseness. Fluency gauges the quality of generated captions in terms of language, accuracy assesses whether the main objects and actions in the caption accurately reflect the picture, and conciseness evaluates the extent to which generative captions are informative and succinct, avoiding unnecessary details.

To ensure evaluation quality, we randomly sampled 10 pictures from the COCO dataset and generated corresponding captions for SCD-Net, BLIP\footnote{For BLIP, we utilized the following page for convenient inference: https://replicate.com/salesforce/blip.}, and our \modelname model. Subsequently, we shuffled the three captions and required volunteers to rate them. To guarantee the reliability of the evaluation, we randomly selected 2 evaluators and calculated their correlation on each metric. This procedure was repeated 5 times, and all results were found to be satisfactory.

As depicted in Tab.~\ref{tab: human_eval}, our model surpasses the previous diffusion-based state-of-the-art SCD-Net in all aspects, achieving comparable results with BLIP. A slight decrease in text quality compared to BLIP may be attributed to the substantial training data used in BLIP's pretraining.
\begin{table} 
\centering
\begin{adjustbox}{max width=1\linewidth}
\begin{tabular}{c|c}
\toprule
       Hyperparameters    &  Values        \\  
       \midrule
       \multicolumn{2}{l}{\textit{Training}} \\
       \midrule
       Batch size & 64*8(GPUs)\\
       Epoch & 60          \\ 
Peak Learning rate & 5e-5 \\
Learning rate schedule & Linear \\
Warmup ratio & 0.1 \\
Optimizer & AdamW \\
$\beta_1$ & 0.9 \\
$\beta_2$ & 0.999 \\
\midrule
       \multicolumn{2}{l}{\textit{Inference}} \\
\midrule
Method & DDIM \\
Sampling Criterion & Minimum Bayes Risk \\

\midrule
       \multicolumn{2}{l}{\textit{Diffusion Process}} \\
\midrule
Diffusion steps & 1000 \\
$\beta$ minimum & 0.0001 \\
$\beta$ maximum & 0.02 \\
$\beta$ schedule & Cosine \\
Classifier free probability & 0.1 \\
Classifier free weight & 1 \\
Self-conditioning probability & 0.5 \\
\midrule
       \multicolumn{2}{l}{\textit{Loss}} \\
\midrule
$\lambda$ & 0.2 \\
Loss type & $l_2$\\
\midrule
       \multicolumn{2}{l}{\textit{Image Encoder}} \\
\midrule
Image size & 224\\
Image Encoder & BLIP$\rm{_{base}}$\\
\midrule
       \multicolumn{2}{l}{\textit{Diffuser Module}} \\
\midrule
Sequence length & 24\\
Hidden size & 768 \\
Layers & 12 \\
FFN size & 3072\\
Attention heads & 16 \\

\bottomrule
\end{tabular}
\end{adjustbox}
\caption{More hyperparameters of our \modelname model. }
\label{tab: hyperpara}
\end{table}

\section{More Hyperparameters} \label{sec: app_hyper}

We list more hyperparameters for \modelname model in Tab.~\ref{tab: hyperpara}.

\section{Mathematical Details for Diffusion Models}\label{app:diff_math}
The training flow of the diffusion models is divided into two phases: the forward diffusion process and the backward denoising process. Given a data point sampled from a real data distribution $x_0\sim q(x)$\footnote{We follow the notation and derivation process of https://lilianweng.github.io/posts/2021-07-11-diffusion-models.}, we define a forward diffusion process in which Gaussian noise is incrementally added to the sample, generating a sequence of noisy samples $x_1, ..., x_T$. The noise scales are controlled by a variance schedule $\beta_t \in (0,1)$, and the density is expressed as
$q(x_t|x_{t-1})= \mathcal{N}(x_t; \sqrt{1-\beta_t}x_{t-1}, \beta_t\mathbf I)$.
Based on the reparameterization trick~\citep{Ho2020DenoisingDP}, a nice property of the above process is that we can sample at any arbitrary time step in a closed form: 
\begin{equation*}
\begin{aligned} x_t=&\sqrt{\alpha_t}x_{t-1}+\sqrt{1-\alpha_t}\epsilon_{t-1} \\ =&\sqrt{\alpha_t}(\sqrt{\alpha_{t-1}}x_{t-2}
+\sqrt{1-\alpha_{t-1}}\epsilon_{t-2})\\&+\sqrt{1-\alpha_t}\epsilon_{t-1}\\ =&\sqrt{\alpha_t \alpha_{t-1}}x_{t-2}+(\sqrt{\alpha_t(1-\alpha_{t-1})}\epsilon_{t-2}\\
&+\sqrt{1-\alpha_t}\epsilon_{t-1})\\ =&\sqrt{\alpha_t \alpha_{t-1}}x_{t-2}+\sqrt{1-\alpha_{t}\alpha_{t-1}}\overline{\epsilon}_{t-2} \\ =& \cdots\\ =&\sqrt{\overline{\alpha}_t}x_0+\sqrt{1-\overline{\alpha}_t}\epsilon. \end{aligned}
\end{equation*}
where $\alpha_t=1-\beta_t$ and $\bar{\alpha}_t=\prod_{i=1}^t \alpha_i$. Thus:

\begin{equation} \label{equ:diffuse}
    q(x_t|x_0) = \mathcal{N}(x_t; \sqrt{\bar\alpha_t}x_0, \sqrt{1-\bar\alpha_t}\mathbf I),
\end{equation}
Furthermore, from this equation, it becomes evident that as $T\to \infty$, $x_T$ converges to an isotropic Gaussian distribution, aligning with the initial condition during inference.

However, obtaining the closed form of the reversed process $q(x_{t-1}|x_t)$ is challenging. Notably, if $\beta_t$ is sufficiently small, the posterior will also be Gaussian. In this context, we can train a model $p_{\theta}(x_{t-1}|x_t)$ to approximate these conditional probabilities:

\begin{equation*} \label{equ: reverse}
    p_{\theta}(x_{t-1}|x_t)=\mathcal{N}(x_{t-1}; \mu_{\theta}(x_t, t), \Sigma_{\theta}(x_t, t)), 
\end{equation*}
where $\mu_{\theta}(x_t, t)$ and $\Sigma_{\theta}(x_t, t)$ are parameterized by a denoising network $f_{\theta}$ like U-Net ~\citep{Ronneberger2015UNetCN} or Transformer~\citep{Vaswani2017AttentionIA}. Similar to VAE~\citep{Kingma2013AutoEncodingVB}, we can derive the variational lower bound to optimize the negative log-likelihood of input $x_0$~\citep{DDPM},
:
\begin{equation*}
    \begin{aligned}
    \mathcal L_{\rm vlb}&=  \mathbb{E}_q[\underbrace{D_{\rm KL}(q(x_t|x_0)||p_\theta(x_T))}_{\mathcal L_T}]-\underbrace{\log p_\theta(x_0|x_1)}_{\mathcal L_0} \\    
    &+\mathbb{E}_q[\sum_{t=2}^T\underbrace{D_{\rm KL}(q(x_{t-1}|x_{t}, x_0)||p_\theta( x_{t-1}|x_t))}_{\mathcal L_{t-1}}].
    \end{aligned}
\end{equation*}

 With an additional condition on $x_0$, the posterior of the forward process $q(x_{t-1}|x_t, x_0)$ can be calculated using Bayes theorem. Then in~\citep{DDPM} they derive:
\begin{equation*}\small
\begin{aligned}
L_t&=\mathbb{E}_{x_0,\epsilon}\left[\frac{1}{2||\Sigma_\theta(x_t,t)||_2^2}||\tilde{\mu}_t(x_t,x_0)-\mu_\theta(x_t,t)||^2\right]\\ 
&=\mathbb{E}_{x_0,\epsilon}\left[\frac{1}{2||\Sigma_\theta(x_t,t)||_2^2}||\frac{1}{\sqrt{\overline{a}_t}}(x_t-\frac{\beta_t}{\sqrt{1-\overline{a}_t}}\epsilon_t)\right.\\
&\qquad \left. -\frac{1}{\sqrt{\overline{a}_t}}(x_t-\frac{\beta_t}{\sqrt{1-\overline{a}_t}}\epsilon_\theta(x_t,t))||^2 \right]\\ 
&=\mathbb{E}_{x_0,\epsilon}\left[\frac{\beta_t^2}{2\alpha_t(1-\overline{\alpha}_t)||\Sigma_\theta||_2^2)}||\epsilon_t-\epsilon_\theta(x_t,t)||^2\right]\\ 
&=\mathbb{E}_{x_0,\epsilon}\Bigg[\frac{\beta_t^2}{2\alpha_t(1-\overline{\alpha}_t)||\Sigma_\theta||_2^2)}\times\\
&\qquad  ||\epsilon_t-   \epsilon_\theta(\sqrt{\overline{\alpha}_t}x_0+\sqrt{1-\overline{\alpha}_t}\epsilon_t,t)||^2\Bigg]
    \end{aligned}
\end{equation*}
Removing the coefficients, a much more simple DDPM learning objective can be obtained:
\begin{equation*} \label{equ: simple loss}
    \mathcal L_{\rm simple}=\sum_{t=1}^T\mathbb{E}_q \big[||\epsilon_t(x_t, x_0)-\epsilon_{\theta}(x_t, t)||^2 \big], 
\end{equation*}
where $\epsilon_t$ is the noise added in original data $x_0$. 
Applied to textual data, \citep{difflm} introduces an even simpler architecture to train a network to predict $x_0$ directly, with the loss function defined as $L=||x_0-f_\theta(x_t,t)||$.

During inference, the reverse process commences by sampling noise from a Gaussian distribution $p(x_T)=\mathcal{N}(x_T;\mathbf{0}, \mathbf{I})$ and iteratively denoising it using $p_{\theta}(x_{t-1}|x_t)$ until reaching $x_0$. In DDIM~\citep{ddim}, a general form is derived from Equation~\ref{equ:diffuse}.
\begin{equation*}
    \begin{aligned}
        x_{t-1}&=\sqrt{\overline{\alpha}_{t-1}}x_0+\sqrt{1-\overline{\alpha}_{t-1}}\epsilon_{t-1}\\ &=\sqrt{\overline{\alpha}_{t-1}}x_0+\sqrt{1-\overline{\alpha}_{t-1}-\sigma_t^2}\epsilon_t\\
        &\qquad +\sigma_t \epsilon\\ &=\sqrt{\overline{\alpha}_{t-1}}x_0+\sqrt{1-\overline{\alpha}_{t-1}-\sigma_t^2}\\
        &\qquad(\frac{x_t-\sqrt{\overline{\alpha}_t}x_0}{\sqrt{1-\overline{\alpha}_t}})+\sigma_t \epsilon\\ 
    \end{aligned}
\end{equation*}

\begin{equation*}
\begin{aligned}
     &q_\sigma(x_{t-1}|x_t,x_0)
    =\mathcal{N}(x_{t-1};\sqrt{\overline{\alpha}_{t-1}}x_0+\\
    &\sqrt{1-\overline{\alpha}_{t-1}-\sigma^2_t}(\frac{x_t-\sqrt{\overline{\alpha}_t}x_0}{\sqrt{1-\overline{\alpha}_t}}), \sigma_t^2\mathbf{I}).
\end{aligned}
\end{equation*}
where $\sigma_{t}^2=\eta \tilde{\beta}_t=\eta \frac{1-\overline{\alpha}_{t-1}}{1-\overline{\alpha}_t}\beta_t$, allowing us to adjust $\eta$ as a hyperparameter to control the sampling stochasticity. The special case of $\eta =0$ renders the sampling process deterministic. This model is referred to as the denoising diffusion implicit model (DDIM). It is noteworthy that DDIM shares the same marginal distribution as DDPM. Consequently, during generation, we can sample only a subset of diffusion steps ${\tau_1, \dots, \tau_S}$, and the inference process becomes:

 \begin{equation*}
     \begin{aligned}
         &q_{\sigma, \tau}(\mathbf{x}_{\tau_{i-1}} \vert \mathbf{x}_{\tau_t}, \mathbf{x}_0)
= \mathcal{N}(\mathbf{x}_{\tau_{i-1}}; \sqrt{\bar{\alpha}_{t-1}}\mathbf{x}_0\\
& + \sqrt{1 - \bar{\alpha}_{t-1} - \sigma_t^2} \frac{\mathbf{x}_{\tau_i} - \sqrt{\bar{\alpha}_t}\mathbf{x}_0}{\sqrt{1 - \bar{\alpha}_t}}, \sigma_t^2 \mathbf{I})
     \end{aligned}
 \end{equation*}
which, significantly reduces inference latency.

\end{document}